\documentclass[11pt,a4paper]{article}
\usepackage[hyperref]{acl2021}
\usepackage{times}
\usepackage{latexsym}

\usepackage{microtype}
\usepackage{ifthen}
\usepackage{color}
\usepackage[utf8]{inputenc}
\usepackage{url}
\usepackage{multicol}
\usepackage{tikz}
\usepackage{booktabs}
\usepackage{csquotes}
\usepackage{soul}
\usepackage{float}
\usetikzlibrary{fit}
\usetikzlibrary{patterns}

\usepackage{enumitem}
\usepackage{tabularx}

\usepackage{booktabs}
\usepackage{amssymb}
\usepackage{xspace}
\usepackage{bbm}
\usepackage{amsmath}
\usepackage{todonotes}
\usepackage{url}
\usepackage{comment}
\usepackage{subcaption}
\usepackage[T1]{fontenc}

\usepackage{url}
\usepackage{enumitem}
\usepackage{booktabs}
\usepackage{multirow}
\usepackage{graphicx}
\usepackage{array}
\usepackage{subcaption}
\usepackage{amsmath}
\usepackage{amsfonts}
\usepackage{amsthm,amssymb,amsopn,bm}
\usepackage{color,soul}
\usepackage{verbatim}
\usepackage{caption}
\usepackage{xcolor,colortbl}
\definecolor{green}{rgb}{0.1,0.1,0.1}
\usepackage{tabulary}
\usepackage{footnote}
\usepackage{blindtext}
\usepackage{algorithm}
\usepackage[noend]{algpseudocode}
\usepackage{stmaryrd}
\usepackage{CJKutf8}
\usepackage{arydshln}
\usepackage[OT2, T1]{fontenc}
\usepackage[russian, english]{babel}

\makesavenoteenv{tabular}
\makesavenoteenv{algorithm}
\usepackage{tabularx,booktabs}
\newcolumntype{Y}{>{\centering\arraybackslash}X}

\setlist{leftmargin=8mm}

\usepackage{tikz}

\definecolor{gitred}{HTML}{FDB8C0}
\definecolor{gitgreen}{HTML}{006400}
\definecolor{chocolate}{HTML}{D2691E}
\definecolor{maroon}{HTML}{800000}
\definecolor{indigo}{HTML}{4B0082}
\definecolor{green}{HTML}{008000}
\definecolor{orange}{HTML}{fc8d62}
\definecolor{purple}{HTML}{8da0cb}

\aclfinalcopy 

\title{Challenges in Information-Seeking QA:\\Unanswerable Questions and Paragraph Retrieval}

\author{Akari Asai \\
  University of Washington \\
  \texttt{akari@cs.washington.edu} \\\And
    Eunsol Choi\footnote{The work started while the author is visiting Google AI.} \\
  The University of Texas at Austin  \\
  \texttt{eunsol@cs.utexas.edu} \\}

\date{}

\begin{document}
\maketitle
\begin{abstract}
Recent pretrained language models ``solved'' many reading comprehension benchmarks, where questions are written with the access to the evidence document. However, datasets containing information-seeking queries where evidence documents are provided after the queries are written independently remain challenging. 
We analyze why answering information-seeking queries is more challenging and where their prevalent unanswerabilities arise, on Natural Questions and TyDi QA. 
Our controlled experiments suggest two headrooms -- paragraph selection and answerability prediction, i.e. whether the paired evidence document contains the answer to the query or not. When provided with a gold paragraph and knowing when to abstain from answering, existing models easily outperform a human annotator. 
However, predicting answerability itself remains challenging. We manually annotate 800 unanswerable examples across six languages on what makes them challenging to answer. With this new data, we conduct per-category answerability prediction, revealing issues in the current dataset collection as well as task formulation.
Together, our study points to avenues for future research in information-seeking question answering, both for dataset creation and model development.\footnote{{Our code and annotated data is publicly available at \url{https://github.com/AkariAsai/unanswerable_qa}.}}
\end{abstract}
\section{Introduction}\label{sec:intro}
Addressing the information needs of users by answering their questions can serve a variety of practical applications.
To answer such information-seeking queries -- where users pose a question because they do not know the answer -- in an unconstrained setting is challenging for annotators as they have to exhaustively search over the web.
To reduce annotator burden, the task has been simplified as reading comprehension: annotators are tasked with finding an answer in a single document. 
Recent pretrained language models surpassed estimated human performance~\cite{Liu2019RoBERTaAR,bert} in many reading comprehension datasets such as SQuAD~\cite{Rajpurkar2016SQuAD10} and CoQA~\cite{Reddy2019CoQAAC}, where questions are posed with an answer in mind. However, those state-of-the-art models have difficulty answering information-seeking questions~\cite{kwiatkowski2019natural,Choi2018QuACQ}.

In this work, {we investigate} what makes information-seeking question answering (QA) more challenging, focusing on the Natural Questions~(NQ;~\citealp{kwiatkowski2019natural}) and TyDi QA~\cite{Clark2020TyDiQA} datasets. Our experimental results from four different models over six languages on NQ and TyDi QA show that most of their headroom can be explained by two subproblems: selecting a paragraph that is relevant to a question and deciding whether the paragraph contains an answer. The datasets are annotated at the document level, with dozens of paragraphs, and finding the correct paragraph is nontrivial. When provided with a gold paragraph and an answer type (i.e., if the question is answerable or not), the performance improves significantly (up to 10\% F1 in NQ), surpassing that of a single human annotator. 

After identifying the importance of answerability prediction, in Section~\ref{sec:evaluation}, we compare a question only baseline, state-of-the-art QA models, and human agreement on this task. For comparison, we also evaluate unanswerability prediction in a reading comprehension dataset {including unanswerable questions}~\cite{Rajpurkar2018KnowWY}. 
While all datasets contain a large proportion of unanswerable questions (33-59\%), they differ in how easily models can detect them. This motivates us to further investigate the source of unanswerability.%

To this end, we quantify the sources of unanswerability by {annotating unanswerable questions from NQ and TyDi QA; we first classify unanswerable questions into six categories and then further annotate answers and alternative knowledge sources when we can find the answers to the unanswerable questions.
Despite the difficulty of annotating questions from the web and crowdsourcing bilingual speakers, we annotated 800 examples across six typologically diverse languages.} 
Our analysis shows that why questions are unanswerable differs based on the dataset or language.
We conduct per-category answerability prediction on those annotated data, and found unanswerable questions from some categories are particularly hard to be identified. 
We provide a detailed analysis for alternative sources of an answer beyond Wikipedia. 
Grounded in our analysis, we suggest avenues for future research, both for dataset creation and model development based on the analysis.

Our contributions are summarized as follows:
\begin{itemize}
    \item We provide in-depth analysis on information-seeking QA datasets, namely on Natural Questions and TyDi QA to identify the remaining headrooms. 
    \item We show that answerability prediction and paragraph retrieval remain challenging even for state-of-the-art models through controlled experiments using four different models.
    \item We manually annotate reasons for unanswerability for 800 examples across six languages, and suggest potential improvements for dataset collections and task design. 
\end{itemize}
\section{Background and Datasets}
We first define the terminology used in this paper. In this work, we focus on a reading comprehension setting, where reference documents (context) are given and thus retrieval is unnecessary, unlike open retrieval QA~\cite{chen2021open}.

{\bf Information-seeking QA} datasets contain questions written by a human who wants to know the answer but doesn't know it yet.
In particular, NQ is a collection of English Google Search Engine queries (anonymized) and TyDi QA is a collection of questions authored by native speakers of 11 languages. 
The answers are annotated post hoc by another annotator, who selects a \textbf{paragraph} with sufficient information to answer ({\bf long answer}). Alternatively, the annotator can select ``unanswerable'' if there is no answer on the page, or if the information required to answer the question is spread across more than one paragraph. If they have identified the long answer, then the annotators are tasked to choose the {\bf short answer}, a span or set of spans within the chosen paragraph, if there is any. 
Questions are collected independently from existing documents, so those datasets tend to have limited lexical overlap between questions and context, which is a common artifact in prior reading comprehension datasets~\cite{sugawara-etal-2018-makes}.

{\bf Reading comprehension} datasets such as SQuAD~\cite{Rajpurkar2016SQuAD10}, by contrast, have been created by asking annotators to write question and answer pairs based on a single provided paragraph.
{SQuAD 2.0~\cite{Rajpurkar2018KnowWY} includes unanswerable questions that are written by annotators who try to write confusing questions based on the single paragraph. }

\begin{table}
\footnotesize
\begin{center}
\begin{tabular}{l|r|r|r|r}
\toprule
{\textbf{Data}} & \multicolumn{2}{c|}{\% Answerable} & {\% } & Avg.\\
& Long Only & Short & Un-ans &  \# of P \\\midrule
NQ & 14.7  &35.2 & 50.1 & 131.3 \\  
{TyDi QA} &  5.4 &  34.8 & 59.9 & 41.1 \\ 
 SQuAD 2.0 & -  &   66.6  &  33.4 & 1.0 \\ 
\bottomrule
\end{tabular} 
\end{center}
\vspace{-0.8em}
\caption{Answer type and paragraph number statistics in three datasets' train portions. ``Avg. \# o P'' denotes the average number of the paragraphs included in the reference context per question. About half of the questions are unanswerable (Un-ans); the rest consist of questions with only paragraph-level answers (Long Only) and additional span-level answers (Short).}\vspace{-0.8em}
\label{tab:answer_type_stat}
\end{table}

As shown in Table~\ref{tab:answer_type_stat}, while unanswerable questions are very common in NQ, TyDi QA and SQuAD 2.0, there are some major differences between the first two datasets and the last: First, NQ and TyDi QA unanswerable questions arise naturally, while SQuAD 2.0 unanswerable questions are artificially created by annotators (e.g. changing an entity name). 
Prior work~\cite{kwiatkowski2019natural} suggests that those questions can be identified as such with little reasoning.
Second, while NQ or TyDi QA models have to select the evidence paragraph (long answer) from dozens of paragraphs, SQuAD 2.0 provides a single reference paragraph. {That lengthy context provided in NQ and TyDi QA requires systems to select and focus on relevant information to answer. }
As of January 2021, the best models on NQ or TyDi QA lag behind humans, while several models surpass human performance on SQuAD and SQuAD 2.0.\footnote{\url{https://rajpurkar.github.io/SQuAD-explorer/}}
{In the following sections, we focus on information-seeking QA datasets, investigating how to improve the answer coverage of those questions that are currently labeled as {\it unanswerable} through several controlled experiments and manual analysis.}
\section{QA performances with Gold Answer Type and Gold Paragraph}\label{sec:endtoend}
We quantify how the two aforementioned subproblems in information-seeking QA -- deciding answer type{, also referred to as answer calibrations~\cite{kamath-etal-2020-selective} or answerability prediction,} and finding a paragraph containing the answer -- affect the final QA performance. 
We conduct oracle analysis on existing models given two pieces of key information: {\bf Gold Paragraph} and {\bf Gold Type}.
In the {\bf Gold Paragraph} setting, we provide the {long answer} to limit the answer space. 
In the {\bf Gold Type} setting, a model outputs the final answer following the gold answer type $t_i\in\{\texttt{short},\texttt{long only},\texttt{unanswerable}\}${, {which correspond to the} questions with short answers,\footnote{The short answer is found inside the long answer, so long answer is also provided.} questions with long answers only, and questions without any answers}, respectively. This lifts the burden of answer calibration from the model.

\subsection{Comparison Systems}
\paragraph{QA models.}
For NQ, we use RikiNet~\cite{liu2020rikinet}\footnote{We contacted authors of RikiNet for the prediction files. We appreciate their help.} and ETC~\cite{Ainslie2020ETCEL}.
{These systems are within 3\% of the best-performing systems on the long answer and short answer prediction tasks as of {January 2021}.}
We use the original mBERT~\cite{bert} baseline for TyDi QA. {RikiNet uses an answer type predictor whose predicted scores are used as biases to the predicted long and short answers. 
ETC and mBERT jointly predict short answer spans and answer types, following \citet{alberti2019bert}.}

\paragraph{Human.}
The NQ authors provide upper-bound performance by estimating the performance of a single annotator (Single), and one of the aggregates of 25 annotators (Super). 
Super-annotator performance is considered as an NQ upper bound. See complete distinction in \citet{kwiatkowski2019natural}.

\begin{table}[t!]
\footnotesize
    \centering
    \begin{tabular}{l|rrr|rrr}
\toprule
 & \multicolumn{3}{c|}{Long answer}& \multicolumn{3}{c}{Short answer}\\
 & P & R & F1 & P & R & F1 \\\midrule
RikiNet & 74.3 &76.3 & 75.2&61.4 & 57.3&59.3\\
w/Gold T & 85.2 & 85.2 & 85.2 &64.6 & 64.6 & 64.6\\ \midrule
ETC & 79.7 & 72.2& 75.8 &67.5& 49.9&57.4\\
w/Gold T &84.6 &84.6 & 84.6& 62.5 &62.5&62.5\\
w/Gold P & - & - & - &67.9 &57.7 &62.4 \\
w/Gold T\&P  & - &- & - &68.9&67.6 &68.3\\\midrule
Human &&&&& \\
- Single  &80.4 & 67.6 & 73.4 & 63.4 & 52.6 & 57.5 \\
- Super & 90.0 & 84.6 & 87.2 & 79.1 & 72.6 & 75.7 \\
 \bottomrule
    \end{tabular}
    \caption{Oracle analysis on the dev set for NQ. {``Gold T'' denotes Gold Type, and ``Gold P'' denotes ``Gold Paragraph''. }
    }
    \label{tab:end-to-end-qa-result-nq}
\end{table}
\subsection{Evaluation Metrics}\label{subsec:evaluations}
The final metric of NQ is based on precision, recall and F1 among the examples where more than one annotators select \texttt{NON-NULL} answers and a model predicts a  \texttt{NON-NULL} answer~\cite{kwiatkowski2019natural}, to prevent a model always outputting unanswerable for achieving high scores. 

TyDi QA evaluation is based on recall, precision and byte-level F1 scores among the examples with answer annotations.
The final score is calculated by taking a macro-average score of the results on 11 target languages. 

\subsection{Results}
\begin{table}[t!]
\small
    \centering
    \begin{tabular}{l|rrr|rrr}
\toprule
 & \multicolumn{3}{c|}{Long answer}&  \multicolumn{3}{c}{Short answer}   \\
 & P & R & F1 & P & R & F1  \\ \midrule
mBERT &64.3 & 66.4 & 65.2 & 58.9 & 50.6 & 54.3\\ 
w/ Gold T & 78.5& 78.5&78.5&60.8&60.8&60.8\\
\midrule
Annotator & 84.4 & 74.5 & 79.9 & 70.8& 62.4 & 70.1\\
 \bottomrule
    \end{tabular}\vspace{-5pt}
    \caption{Oracle analysis on the dev set of TyDi QA. {``Gold T'' denotes Gold Type.}}\vspace{-5pt}
    \label{tab:end-to-end-qa-result-tydi}
\end{table}
Table~\ref{tab:end-to-end-qa-result-nq} presents oracle analysis on NQ. Having access to gold answer type and gold paragraph is almost equally crucial for short answer performance on NQ. 
For long answers, we observe that the models rank the paragraphs correctly but struggle to decide when to abstain from answering.
{When the gold type is given, ETC reaches 84.6 F1 for the long answer task, which is only 2.6 points behind the upper bound, and significantly outperforms single annotator performance.}
Provided both gold paragraph and answer type (``Gold T\&P''), the model's short answer F1 score reaches 10\% above that of a single annotator, while slightly behind super human performance. 
For short answers, providing gold paragraph can improve ETC's performance by 5 points, gaining mostly in recall. Having the gold answer type information also significantly improves recall at a small cost of precision.

Table~\ref{tab:end-to-end-qa-result-tydi} shows that a similar pattern holds in TyDi QA: answerability prediction is a remaining challenge for TyDi QA model.\footnote{We do not experiment with Gold P setting for TyDi QA, as it's included in the original paper~\cite{Clark2020TyDiQA}.} 
Given the gold type information, the long answer F1 score is only 1.4 points below the human performance. {These results suggest that our models performed well when selecting plausible answers and would benefit from improved answerability prediction.
}
\label{sec:oracle_qa_performance}
\section{Answerability Prediction}
\label{sec:evaluation}
We first quantitatively analyze how easy it is to estimate answerability from the question alone, and then we test the state-of-the-art models' performance to see how well our complex models {given question and the gold context} perform on this task. 
We conduct the same experiments on SQuAD 2.0, to highlight the unique challenges of the information-seeking queries.

Each example consists of a question $q_i$, a list of paragraphs of an evidence document $d_i$, and a list of answer annotations $A_i$, which are aggregated into an answer type $t_i\in\{\texttt{short},\texttt{long},\texttt{unanswerable}\}$.

\subsection{Models}\label{subsec:models}
\paragraph{Majority baseline.}
We output the most frequent label for each dataset (i.e., \texttt{short} for NQ, \texttt{unanswerable} for TyDi QA and SQuAD 2.0).

\paragraph{Question only model (Q only).}
This model takes a question and classify it into one of {three classes (i.e., \texttt{short},\texttt{long},\texttt{unanswerable})} solely based on the question input. In particular, we use a BERT-based classifier: encode each input question with BERT, and use the $\textsc{[CLS]}$ token as the summary representation to classify. Experimental details can be found in the appendix.

\paragraph{QA models.}
{We convert the state-of-the-art QA models' final predictions into answer type predictions. When a QA system outputs any short/long answers, we map them to \texttt{short} / \texttt{long} type; otherwise we map them to \texttt{unanswerable}. }
We use ETC for NQ, and mBERT baseline for TyDi QA as in Section~\ref{sec:oracle_qa_performance}. For SQuAD 2.0, we use Retro-reader~\cite{zhang2020retrospective}.\footnote{We contacted authors of Retro-reader for the prediction file. We appreciate their help.} 
The evaluation script of NQ and TyDi QA calibrates the answer type for each question by thresholding long and short answers respectively to optimize the F1 score. We use the final predictions after this calibration process.

\paragraph{Human.}
We compare the models' performance with two types of human performance: binary and aggregate. 
``Binary'' evaluation computes pair-wise agreements among all combinations of 5 annotators for NQ and 3 annotators for TyDi QA.
``Aggregate'' evaluation compares each annotator's label to the majority label selected by the annotators. 
This inflates human performance modestly as each annotator's own label contributes to the consensus label.

\begin{table}[t!]
\footnotesize
    \centering
    \begin{tabular}{l|r|r|r|r|r}
\toprule
\multirow{2}{*}{Model} &  \multicolumn{2}{c|}{NQ (ETC)} & \multicolumn{2}{c|}{TyDi} & \scriptsize{SQuAD} \\
 & 3-way & 2-way &3-way&2-way&2-way\\\midrule
Majority   & 50.9 & 58.9 & 58.2 & 58.2 & 50.0 \\
Q Only &  65.5 & 72.7 & 69.8 & 70.2 & 63.0\\
QA Model  &72.0 & 82.5 & 74.2 & 79.4 &  94.1 \\ 
\midrule
Human & &&&&\\
 - binary &71.0  & 78.9 & 88.1 &  86.9 & - \\
 - aggregate &79.6 & 85.6 & 93.3 & 94.0 & -  \\
 \bottomrule
    \end{tabular}\vspace{-4pt}
    \caption{Answer type classification accuracy: $\texttt{long}, \texttt{short}, \texttt{none}$ for three-way classification and {answerable},{unanswerable} for two-way classification.}\vspace{-6pt}
    \label{tab:baseline_performance}
\end{table}

\subsection{Results}
The results in Table~\ref{tab:baseline_performance} indicate the different characteristics of the naturally occurring and artificially annotated unanswerable questions. 
Question only models yield over 70\% accuracy in NQ and TyDi QA, showing there are clues in the question alone, as suggested in ~\citet{liu2020rikinet}. 
While models often outperform binary agreement score between two annotators, the answer type prediction component of ETC performs on par with the Q only model, suggesting that answerability calibration happens mainly at the F1 optimization processing. 

\paragraph{Which unanswerable questions can be easily identified?}
We randomly sample 50 NQ examples which both Q only and ETC successfully answered. 32\% of them are obviously too vague or are not valid questions (e.g., ``bye and bye going to see the king by blind willie johnson'', ``history of 1st world war in Bangla language''). 13\% of them include keywords that are likely to make the questions unanswerable (e.g., ``which of {\bf the following} would result in an snp?''). 14\% of the questions require complex reasoning, in particular, listing entities or finding a maximum / best one (e.g., ``top 10 air defense systems in the world''), which are often annotated as unanswerable in NQ due to the difficulty of finding a single paragraph answering the questions. Models, including the Q only models, seem to easily recognize such questions.

\paragraph{Comparison with SQuAD 2.0.}
In SQuAD 2.0, somewhat surprisingly, the question only baseline achieved only 63\% accuracy. 
We hypothesize that crowdworkers successfully generated unanswerable questions that largely resemble answerable questions, which prevents the question only model from exploiting artifacts in question surface forms. 
However, when the context was provided, the QA model achieves almost 95\% accuracy, indicating that detecting unanswerability becomes substantially easier when the correct context is given. 
{\citet{Yatskar2019AQC} finds the unanswerable questions in SQuAD 2.0 focus on simulating questioner confusion (e.g., adding made-up entities, introducing contradicting facts, topic error)}, which the current state-of-the-art models can recognize when the short reference context is given. By design, these questions are clearly unanswerable, unlike information-seeking queries which can be partially answerable. Thus, identifying unanswerable information-seeking queries poses additional challenges beyond matching questions and contexts.
\label{sec:annotation}
\begin{table*}
\small
\begin{center}
\begin{tabular}{l|l|p{140pt}|p{55pt}|p{35pt}}
\toprule
\textbf{Type} & \textbf{Sub-Type} & Query & Wiki Page Title & Answer\\ \midrule
 \multirow{3}{*}{Retrieval Miss}& factoid question ({\bf Fact}) &when is this is us season 2 released on dvd & This Is Us (season 2)& September 11, 2018\\ \cline{2-5}
 & non-factoid question ({\bf Non-F}) &what is the difference between a bernese mountain dog and a swiss mountain dog & Bernese Mountain Dog & - \\ \cline{2-5}
 & {multi-evidence} question ({\bf Multi}) &{how many states in india have at least one international border} & Border \\\midrule
    \multirow{2}{*}{Invalid QA} & Invalid questions ({\bf q.}) &the judds love can build a bridge album & Love Can Build a Bridge (album)\\ \cline{2-5}
 & false premise ({\bf false}) & what harry potter movie came out in 2008 & Harry Potter (film series) & - \\\cline{2-5}
 & Invalid answers ({\bf ans.}) &who played will smith's girlfriend in independence day& Independence Day (1996 film) & Vivica A. Fox \\
 \bottomrule
\end{tabular} 
\end{center}\vspace{-1em}
\caption{Types of unanswerable questions and their examples in NQ.}\vspace{-1em}
\label{tab:unanswerable_examples}
\end{table*}

\section{Annotating Unanswerability}\label{sec:manual}
In this section, we conduct an in-depth analysis to answer the following questions: (i) where the unanswerability in information-seeking QA arises, (ii) whether we can answer those unanswerable questions when we have access to more knowledge sources beyond a single provided Wikipedia article, and (iii) what kinds of questions remain unanswerable when these steps are taken. 
To this end, we annotate {800} unanswerable questions from NQ and TyDi QA across six languages. Then, we conduct per-category performance analysis to determine the types of questions for which our models fail to predict answerability.

\subsection{Categories of Unanswerable Questions}
We first define the categories of the unanswerable questions. 
\noindent {\bf Retrieval miss}\hspace{1pt} includes questions that are valid and answerable, but paired with a document which does not contain a \textit{single} paragraph which can answer the question. 
We subdivide this category into three categories based on the question types: {\bf factoid}, {\bf non-factoid}, and {\bf multi-evidence} questions. 
\textbf{Factoid} questions are unanswerable due to the failure of retrieving articles with answers available on the web. These questions fall into two categories: where the Wikipedia documents including answers are not retrieved by Google Search, or where Wikipedia does not contain articles answering the questions so alternative knowledge sources (e.g., non-Wikipedia articles) are necessary.
We also find a small number of examples whose answers cannot be found on the web even when we exhaustively searched dozens of web-pages.\footnote{Such cases were more common in low resource languages.}
\textbf{Non-factoid} questions cover complex queries {whose} answers are often longer than a single sentence and {no single paragraphs fully address the questions}. 
Lastly, \textbf{multi-evidence} questions require reasoning over multiple facts such as multi-hop questions~\cite{Yang2018HotpotQAAD,Dua2019DROPAR}. {A question is assigned this category only when the authors need to combine information scattered in two or more paragraphs or articles. Theoretically, the boundaries among the categories can overlap (i.e., there could be one paragraph that concisely answers the query, which we fail to retrieve), but in practice, we achieved a reasonable annotation agreement. 
}

{\bf Invalid QA} includes {\bf invalid questions}, {\bf false premise} and {\bf invalid answers}. 
\textbf{Invalid questions} are ill-defined queries, where we can only vaguely guess the questioner's intent.
NQ authors found 14\% of NQ questions are marked as bad questions; here, we focus on the {\it unanswerable} subset of the original data. {We regard queries with too much ambiguity or subjectivity to determine single answers as invalid questions (e.g., where is turkey commodity largely produced in {\it our country})}.
{{\bf False premise}~\cite{Kim2021WhichLI} are questions based on incorrect presuppositions. For example, the question in Table~\ref{tab:unanswerable_examples} is valid, but no Harry Potter movie was released in 2008, as its sixth movie release was pushed back from 2008 to 2009 to booster its release schedule.}
{\bf Invalid answers} are annotation errors, where the annotator missed an answer existing in the provided evidence document.

\subsection{Manual Study Setting}
We randomly sampled and intensively annotated a total of 450 unanswerable questions from the NQ development set, and 350 unanswerable questions across five languages from the TyDi QA development set. 
Here, we sample questions where annotators unanimously agreed that no answer exists. 
See Table~\ref{tab:data_analysis_main} for the statistics. 
For NQ, the authors {of this paper} annotated 100 examples and adjudicated the annotations to clarify common confusions. The remaining 350 questions were annotated individually. 
Before the adjudication, the annotators agreed on roughly 70\% of the questions. After this adjudication process, the agreements on new samples reached over 90\%.

For TyDi QA, we recruit five native speakers to annotate examples in Bengali, Japanese, Korean, Russian, and Telugu. We provide detailed instructions given the adjudication process, and closely communicate with each annotator when they experienced difficulty deciding among multiple categories. Similar to NQ annotation, annotators searched the answers using Google Search, in both the target language and English, referring to any web pages (not limited to Wikipedia) and re-annotated the answer, while classifying questions into the categories described earlier.

\subsection{Results}
\paragraph{Causes of unanswerability.}
Table~\ref{tab:data_analysis_main} summarizes our manual analysis. 
We found different patterns of unanswerability in the two datasets. 
Invalid answers were relatively rare in both, which shows they are high quality. 
We observe that invalid answers are more common for questions where annotators need to skim through large reference documents.
In NQ, where the questions are naturally collected from user queries, ill-defined queries were prevalent (such queries account for 14\% of the whole NQ data, but 38\% of the {\it unanswerable} subset). In TyDi QA, document retrieval was a major issue across all five languages (50-74\%), and a significantly larger proportion of re-annotated answers were found in other Wikipedia pages (50\% in TyDi QA v.s. 21.8\% in NQ), indicating that the retrieval system used for document selection made more mistakes. 
Document retrieval is a crucial part of QA, not just for modeling but also for dataset construction. 
We observe more complex and challenging questions in some TyDi QA languages; 20\% of the unanswerable questions in Korean and 32\% of the unanswerable questions in Russian require multiple paragraphs to answer, as opposed to 6\% in NQ. 
\begin{table}[t!]
\footnotesize
\begin{center}
\begin{tabular}{l|r|r|r|r|r|r|r}
\toprule
&  &  \multicolumn{3}{|c|}{\textbf{\% Retrieval Miss}}& \multicolumn{3}{|c}{\textbf{\% Invalid }}  \\ 
&   {\bf N} & Fact & Non-F & Multi & q. & false &ans. \\\midrule
NQ &450  & 25 & 20 & 6 & 38 & 3 & 8 \\  \midrule
Bn & 50 & 68 & 0  & 4 &  4 &6& 18 \\
Ja & 100 & 61 & 11 & 15 & 2 & 4 & 7\\
Ko & 100&57 & 8 & 20  & 14 & 0 & 1\\ 
Ru & 50 & 50 & 6 & 32 & 8 & 0 &4 \\
Te & 50 & 74 & 2 & 0& 14 & 0 & 12  \\
\bottomrule
\end{tabular} 
\end{center}\vspace{-1em}
\caption{The manual classification results based on the unanswerable question categories (Table~\ref{tab:unanswerable_examples}) on N examples per row. 
{The bottom five rows represent TyDi QA Bengali, Japanese, Korean, Russian and Telugu, respectively in order. }
}
\label{tab:data_analysis_main}
\end{table}
\begin{table}[t!]
\footnotesize
    \centering
    \begin{tabular}{l|r|r|r|r}
\toprule
& & \multicolumn{3}{c}{ {\bf Number (\%)}} \\
dataset &total & Ib / tab & diff. Wiki& non-Wiki \\\midrule
NQ  & 119 & 3 (2.5) & 26 (21.8) & 119 (75.6) \\\hline
Bn & 40 & 9  (22.5)& 27 (67.5) &  4 (10.0)\\
Ja  & 60 & 10 (16.7) & 40 (66.7)& 10 (16.7) \\
Ko &54 &  13 (24.1) & 30 (55.6) &11 (20.3) \\
Ru &23 &10 (43.4) & 8 (34.8)& 5 (21.7) \\
Te & 23 & 4 (17.4) & 5 (21.8) &14 (60.9) \\
 \bottomrule
    \end{tabular}
    \caption{The knowledge sources for retrieval miss questions in NQ and TyDi Bengali, Japanese, Korean, Russian and Telugu annotation. {The bottom five rows represent TyDi QA Bengali, Japanese, Korean, Russian and Telugu, respectively. ``Ib / tab'' denotes infobox or table in the same Wikipedia pages, ``diff. Wiki'' denotes different Wikipedia pages, and ``non-Wiki'' denotes non-Wikipedia webpages. } }
    \label{tab:breakdown_new_annotation}
\end{table}
\begin{table*}[ht!]
\small
\begin{center}
\begin{tabular}{l|p{140pt}|p{70pt}|p{70pt}|p{20pt}}
\toprule
\textbf{Sub-Type} &   \multicolumn{3}{c}{\textbf{Example}}  \\ 
 & Query & Original Wiki Title & New article & Answer\\ \midrule
different Wikipedia &\begin{CJK}{UTF8}{min}ヴィンセント・トマス・ロンバルディは何歳で死去した？\end{CJK} (At what age did Vincent Thomas Lombardy die?) & \begin{CJK}{UTF8}{min}森川智之\end{CJK} (Toshiyuki Morikawa) & \begin{CJK}{UTF8}{min}ヴィンス・ロンバルディ\end{CJK} (Vince Lombardi) & 57 \\ \cline{2-5}
&\foreignlanguage{russian}{Сколько марок было выпущено в СССР в 1938?} (How many stamps were produced in the USSR in 1938?) & \foreignlanguage{russian}{ Почтовые марки СССР} (Postage stamps of the USSR) & \foreignlanguage{russian}{Знаки почтовой оплаты СССР (1938)} (Signs of the postage of the USSR (1938)) & 97 \\\hline
not Wikipedia &\begin{CJK}{UTF8}{mj}포켓몬스터에서 가장 큰 포켓몬은 무엇인가?\end{CJK} (What's the largest pokemon in pokemonster?) & \begin{CJK}{UTF8}{mj}하야시바라 메구미\end{CJK} (Hayashibara Megumi) & Top 10 Largest Pokémon & Onix\\ \cline{2-5}
&\begin{CJK}{UTF8}{min}日本で平均的に車が買い替えられる頻度は？\end{CJK} (How often do people buy a new car on average in Japan?) & \begin{CJK}{UTF8}{min}モータリゼーション\end{CJK} (Effects of the car on societies) & \begin{CJK}{UTF8}{min}2017年度乗用車市場動向調査\end{CJK} (FY 2017 Private Car Market Survey) & 7.0 \\
 \bottomrule
\end{tabular} 
\end{center}\vspace*{-10pt}
\caption{Examples of retrieval miss (factoid) questions in TyDi Japanese, Korean and Russian subsets. English translations annotated by native speakers are written in the parentheses.}
\label{tab:unanswerable_example_retrieval_miss}
\end{table*}
\begin{table*}
\small
\begin{center}
\begin{tabular}{p{50pt}|p{240pt}|p{120pt}}
\toprule
\textbf{Sub-Type} &   \multicolumn{2}{c}{\textbf{Example}}  \\ 
 & Query & Wiki Page Title\\ \midrule
non-factoid question & \begin{CJK}{UTF8}{mj}공리주의는 영국에 어떤 영향을 미쳤는가?\end{CJK} (How did utilitarianism affect UK?) & \begin{CJK}{UTF8}{mj}제러미 벤담\end{CJK}~(Jeremy Bentham) \\ \cline{2-3}
 & \foreignlanguage{russian}{Почему надо поджигать абсент?} (Why should you lit absinthe on fire?) & \foreignlanguage{russian}{Абсент}~(Absinthe)  \\ \cline{2-3}
& \begin{CJK}{UTF8}{min}スペースシャトルと宇宙船の違いは何？\end{CJK} (What is the difference between a space shuttle and a spaceship?) & \begin{CJK}{UTF8}{min}宇宙船\end{CJK}~(Space ship) \\ 
\hline
{multi-evidence} question &\begin{CJK}{UTF8}{mj}닥터 후 시리즈 중 가장 높은  시청률을 기록한 시리즈는 무엇인가? \end{CJK} (Which Doctor Who series scored the highest view rate?) & \begin{CJK}{UTF8}{mj}코드 블루 -닥터헬기긴급구명-\end{CJK} (Code Blue (TV series)) \\\cline{2-3} &\begin{CJK}{UTF8}{min}進化論裁判はアメリカ以外で起きたことはある？ \end{CJK} (Has any legal case about Creation and evolution in public education ever happened outside of the US?) & \begin{CJK}{UTF8}{min}進化論\end{CJK} (Darwinism) \\
 \bottomrule
\end{tabular} 
\end{center}\vspace*{-10pt}
\caption{Examples of non-factoid and multi-evidence questions in TyDi Japanese, Korean and Russian subsets.}
\label{tab:unanswerable_examples_complex}
\end{table*}
\paragraph{Alternative knowledge sources.}

{
Table~\ref{tab:breakdown_new_annotation} shows the breakdown of the newly annotated answer sources for the ``retrieval miss (factoid)'' questions. 
As mentioned above, in TyDi QA new answers are found in other Wikipedia pages (66.7\% of retrieval miss in Japanese subset, 55.6\% in Korean subset and 34.8\% in Russian), while in NQ, the majority of the answers are from non-Wikipedia websites, which indicates that using Wikipedia as the single knowledge source hurts the coverage of answerability.
{Table~\ref{tab:unanswerable_example_retrieval_miss} shows retrieval miss (factoid) questions in TyDi Japanese, Korean and Russian subsets. In the first example, the retrieved document is about a voice actor who has acted on a character named Vincent. Yet, Japanese Wikipedia has an article about Vince Lombardi, and we could find the correct answer ``57'' there. 
The second group shows two examples where we cannot have Wikipedia articles with sufficient information to answer but can find non-Wikipedia articles on the web. For example, we cannot find useful Korean Wikipedia articles for a question about Pokemon, but a non-Wikipedia Pokemon fandom page clearly answers this question. This is also prevalent in NQ. We provide a list of the alternative web articles sampled from the retrieval misses (factoid) cases of NQ in Table~\ref{tab:alternative_sources} in the appendix. }

For the TyDi QA dataset, answers were sometimes found in tables or infoboxes of provided Wikipedia documents. 
This is because TyDi QA removes non-paragraph elements (e.g., Table, List, Infobox) to focus on the modeling challenges of multilingual text~\cite{Clark2020TyDiQA}.  
WikiData also provides an alternative source of information, covering roughly 15\% of queries.
These results show the potential of searching heterogeneous knowledge sources~\cite{Chen2020HybridQAAD,oguz2020unified} to increase answer coverage. Alternatively, \citet{xorqa} show that searching documents in another language significantly increases the answer coverage of the questions particularly in low-resource languages.
Lastly, a non-negligible number of Telugu and Bengali questions cannot be answered even after an extensive search over multiple documents due to the lack of information on the web. 
A Bengali question asks ``Who is the father of famous space researcher Abdus Sattar Khan (a Bangladeshi scientist)?'', and our annotator could not find any supporting documents for this question.
}

\paragraph{Limitations of the current task designs.}
Table~\ref{tab:unanswerable_examples_complex} shows non-factoid or multi-evidence questions from TyDi QA, which are marked as unanswerable partially due to the task formulation -- answers have to be {\it extracted} from a single paragraph based on the information provided in the evidence document. 
On the first three examples of non-factoid questions, we have found that to completely answer the questions, we need to combine evidence from multiple paragraphs and to write descriptive answers. 
The second group shows several examples for multi-evidence questions. Although they are not typical compositional questions in multi-hop QA datasets~\cite{Yang2018HotpotQAAD}, it requires comparison across several entities. 

\subsection{Per-category Performance}
How challenging is it to detect unanswerablity from different causes?
Table~\ref{tab:per_caetgory_prediction} shows the per-category performance of answerability prediction using the models from Section~\ref{sec:evaluation}. 
Both Q only and QA models show the lowest error rate on invalid questions on NQ, suggesting that those questions can be easily predicted as unanswerable, even from the question surface only. 
Unsurprisingly, all models struggle on the invalid answer category. 
We found that in some of those cases, our model finds the correct answers but is penalized.
Detecting factoid questions’ unanswerability is harder when reference documents are incorrect but look relevant due to some lexical overlap to the questions. For example, given a question ``who sang the song angel of my life'' and the paired document saying ``My Life is a song by Billy Joel that first appeared on his 1978'', which is about a different song, our QA model extracts Billy Joel as the answer with a high confidence score. 
This shows that even the state-of-the-art models can be fooled by lexical overlap. 

\begin{table}[ht!]
\small
\begin{center}
\begin{tabular}{l|rrr|rr}
\toprule
 & \multicolumn{3}{c}{NQ} & \multicolumn{2}{c}{TyDi (MBERT)} \\
category & Q only  & QA & \# ex. & Q only & \# ex. \\\midrule 
Fact &33.9 & 24.1 & 112 & 21.2 & 212 \\
Non-F & 16.9 & 22.9 &  87 & 17.4 & 23 \\
Multi  & 27.5 & 18.5 & 29 & 17.0 & 53 \\\hline
q. & 8.5 & 7.2 & 165 & 20.6 & 29 \\
false & 42.8 &  14.3 & 14 &  14.3 & 7 \\
ans.  & 47.2 &  48.6 & 35 & 32.0 & 25\\
\bottomrule
\end{tabular} 
\end{center}\vspace{-1em}
\caption{{Per-category answerablity prediction error rates. The categories correspond to the six categories in Table~~\ref{tab:unanswerable_examples} and `\# ex' column represents the number of examples in each category. 
}
}
\label{tab:per_caetgory_prediction}
\end{table}

\subsection{Discussion}
We summarize directions for future work from the manual analysis. First, going beyond Wikipedia as the only source of information is effective to increase the answer coverage. Many of the unanswerable questions in NQ or TyDi QA can be answered if we use non-Wikipedia web pages {(e.g., IMDb)} or structured knowledge bases (e.g., WikiData). 
Alternative web pages where we have found answers have diverse formats and writing styles. 
Searching those documents to answer information-seeking QA may introduce additional modeling challenges such as domain adaptation or generalization. To our knowledge, there is no existing large-scale dataset addressing this topic. 
Although there are several new reading comprehension datasets focusing on reasoning across multiple modalities~\cite{talmor2021multimodalqa,hannan2020manymodalqa}, {limited prior work integrate heterogeneous knowledge sources for open-domain or information-seeking QA~\cite{oguz2020unified,chen2021open}. }

Invalid or ambiguous queries are common in information-seeking QA, where questions are often under-specified. We observed there are many ambiguous questions included in NQ data. 
Consistent with the findings of \citet{Min2020AmbigQAAA}, we have found that many of the ambiguous questions or ill-posed questions can be fixed by small edits, and we suggest asking annotators to edit those questions or asking them a follow-up clarification instead of simply marking and leaving the questions as is in the future information-seeking QA dataset creation.  

Lastly, we argue that the common task formulation, extracting a span or a paragraph from a single document, limits answer coverage. 
To further improve, models should be allowed to generate the answer based on the evidence document~\cite{Lewis2020RetrievalAugmentedGF}, instead of limiting to selecting a single span in the document. Evaluating the correctness of free-form answers is more challenging, and requires further research~\cite{Chen2020MOCHAAD}.

While all the individual pieces might be revealed in independent studies~\cite{Min2020AmbigQAAA,oguz2020unified}, our study quantifies how much each factor accounts for reducing answer coverage. 
\section{Related Work}
\paragraph{Analyzing unanswerable questions.}
There is prior work that seeks to understand unanswerability in reading comprehension datasets. 
\citet{Yatskar2019AQC} analyzes unanswerable questions in SQuAD 2.0 and two conversational reading comprehension datasets, namely CoQA and QuAC, while we focus on information-seeking QA datasets to understand the potential dataset collection improvements and quantify the modeling challenges of the state-of-the-art QA models.

\citet{ravichander-etal-2019-question} compare unanswerable factors between NQ and a QA dataset on privacy policies.
This work primarily focuses on a privacy QA, which leads to differences of the categorizations of the unanswerable questions. 
We search alternative knowledge sources as well as the answers to understand how we could improve answer coverage from dataset creation perspective and connect the annotation results with answerability prediction experiments for modeling improvements.

\paragraph{Answer Calibrations.}
Answerability prediction can bring practical values, when errors are expensive but abstaining from it is less so~\cite{kamath-etal-2020-selective}.
While predicting answerability has been studied in SQuAD 2.0~\cite{zhang2020retrospective,hu2019read+}, the unanswerability in SQuAD 2.0 has different characteristics from unanswerability in information-seeking QA as we discussed above.
To handle unanswerable questions in information-seeking QA, models either adopt threshold based answerable verification~\cite{bert}, or introduce an extra layer to classify unanswerablity and training the model jointly~\cite{zhang2019sg,yang2019xlnet}. 
\citet{kamath-etal-2020-selective} observes the difficulty of answer calibrations, especially under domain shift. 

\paragraph{Artifacts in datasets.}
Recent work~\cite{Gururangan2018AnnotationAI,kaushik-lipton-2018-much,sugawara-etal-2018-makes,Chen2019UnderstandingDD} exhibited that models can capture annotation bias in crowdsourced data effectively, achieving high performance when only provided with a partial input. 
Although NQ and TyDi QA attempt to avoid such typical artifacts of QA data by annotating questions independently from the existing documents~\cite{Clark2020TyDiQA}, we found artifacts in question surface forms can let models easily predict answerability with a partial input (i.e., question only).

\section{Conclusion}
We provide the first in-depth analysis on information-seeking QA datasets to inspect where unanswerability arises and quantify the remaining modeling challenges.
Our controlled experiments identifies two remaining headrooms, answerability prediction and paragraph selection. 
Observing a large percentage of questions are unanswerable, we provide manual analysis studying \textit{why} questions are unanswerable and make suggestions to improve answer coverage: (1) going beyond Wikipedia {textual information} as the only source of information, (2) addressing ambiguous queries {instead of simply marking and leaving the questions as is}, (3) enable accessing multiple documents and introducing abstractive answers for non-factoid questions. 
Together, our work shed light on future work for information-seeking QA, both for modeling and dataset design.

\section*{Legal and Ethical Considerations}
{
All of the manual annotations conducted by the authors of the papers and our collaborators. The NQ and TyDi QA data is publicly available and further analysis built upon on them is indeed encouraged. 
This work would encourage future dataset creation and model development for information-seeking QA towards building a QA model that could work well on users' actual queries. 
}
\section*{Acknowledgments}
We thank Jon Clark, Michael Collins, Kenton Lee, Tom Kwiatkowski, Jennimaria Palomaki, Sewon Min, Colin Lockard, David Wadden, Yizhong Wang for helpful feedback and discussion. 
We thank Vitaly Nikolaev for helping with the Russian data annotation, Trina Chatterjee for help with Bengali data annotation, and for Aditya Kusupati for Telegu data annotation. We also thank the authors of RikiNet, Retro-reader and ETC for their cooperation on analyzing their system outputs.{
We are grateful for the feedback and suggestions from the anonymous reviewers. This research was supported by gifts from Google and the Nakajima Foundation Fellowship.
}
\newpage
\bibliographystyle{acl_natbib}
\bibliography{acl2021}
\clearpage
\appendix
\section{Annotation Instruction}
The authors annotated examples in the following process.
\begin{itemize}
    \item (Step 1) Translate the query, if not in English.
    \item (Step 2) Decide whether the query is valid $\rightarrow$ if not, mark as (5) Annotation Error (Question is ambiguous or unclear), if not, go to step 3.
    \item (Step 3) If the query is valid, look at the linked document. if the answer is in the document, write down the answer in the ``answer'' column of the spreadsheet, mark it as (4) Invalid QA.  The corner case here is if the answer is in the infobox, according to TyDi definition it won't work. so in this case mark as (1) Retrieval Error (Factoid question) and label as "Type of missing information: no description in paragraphs, but can be answered based on infobox or table". If you cannot find the answer in the document, go to step 4. 
    \item (Step 4) If the answer is not in the document, google question to find an answer. 
    - If there's a factoid answer found, mark it as (1) Retrieval Error (factoid question) and copy-paste the answer. Mark the source of the answer -- whether from other Wikipedia page, or in English Wikipedia, or in the web. If the answer is non factoid and can be found, mark it as (2) Retrieval Error (non-factoid question), and copy paste a link where the answer. Mark the source of the answer -- whether from another Wikipedia page, or in the web. 
    - If the question is very complex and basically you can't find an answer, mark it as (3) Retrieval Error (complex question). 
\end{itemize}

\section{Experimental Details of Question Only baseline}
Our implementations are all based on PyTorch. 
In particular, to implement our
classification based and span-based model, we use
\texttt{pytorch-transformers}~\cite{wolf2019transformers}.\footnote{\url{https://github.com/huggingface/transformers}}
We use \texttt{bert-base-uncased} model for NQ and SQuAD, \texttt{bert-base-multilingual-uncased} for TyDi as initial pre-trained models. The training batch size is set to 8, the learning rate is set to 5e-5. We set the maximum total input sequence length to 128. We train our model with a single GeForce RTX 2080 with 12 GB memory for three epochs, which roughly takes around 15 minutes, 30 minutes and 45 minutes for each epochs on SQuAD 2.0, TyDi and NQ, respectively. 
The hyperparameters are manually searched by authors, and we use the same hyperparameters across datasets that perform best on NQ Q-only experiments.

\section{Additional Annotation Results}
\subsection{Examples of alternative Web pages for NQ Retrieval Miss (Factoid)}
Table~\ref{tab:alternative_sources} shows several examples of alternative web pages where we could find answers to originally {\it unanswerable} questions. Although those additional knowledge sources are highly useful, they are diverse (from a fandom site to a shopping web site), and all have different formats and writing styles. 

\begin{table*}
\small
\begin{center}
\begin{tabular}{p{70pt}|p{130pt}|p{150pt}|p{40pt}}
\toprule
Query & New article & Paragraph & Answer\\ \midrule
when is fairy tail ep 278 coming out & Fairy Tail Wiki at: \url{https://fairytail.fandom.com/} & Information
Japan Air Date October 7, 2018 & October 7, 2018 \\\hline
where was the American horror story cult filmed & American Horror Story ``Cult'' Fillming Locations at: \url{https://hollywoodfilminglocations.com/} &The horror show ``American Horror Story Cult'' starring Sarah Paulson \&Ryan Murphy was filmed on location throughout Southern California and Michigan.  & Southern California and Michigan \\\hline
what are the main types of meat eaten in the uk & A Roundup Of The Most Popular Meats Eaten In The UK at: \url{https://newyorkstreetfood.com/} & Beef (33\% out of 94\% consider beef as their top choice):
Beef is the most preferred choice among British people  & beef \\\hline
around the world in 80 days book pages & Around the World in 80 Days Paperback – November 6, 2018 at: \url{https://www.amazon.com/} & Publisher : CreateSpace Independent Publishing Platform (November 6, 2018) Language : English Paperback : 130 pages & 130 pages \\
 \bottomrule
\end{tabular} 
\end{center}\vspace*{-10pt}
\caption{Examples of the alternative websites we could find answers to the retrieval miss (factoid) questions from Natural Questions.}
\label{tab:alternative_sources}
\end{table*}

\subsection{Examples of retrieval misses without any alternative knowledge sources}
Table~\ref{tab:missing_info} shows the examples where we cannot find any alternative knowledge sources on the web. Those questions often ask some entities who are not widely known but are closely related to certain culture or community (e.g., a Japanese athlete, geography of an Indian village).

\begin{table*}
\small
\begin{center}
\begin{tabular}{p{70pt}|p{300pt}}
\toprule
Dataset (language) &Query \\ \midrule
TyDi (Telugu) & What is the main agricultural crop in Onuru village (a village in India)? \\
TyDi (Telugu) & As of 2002, what is the biggest construction in Tenali town (a city of India)? \\
TyDi (Japanese) & What is Yuta Shitara (a Japanese long-distance runner.)'s best record for 10000 meters? \\
NQ (English) & how many blocks does hassan whiteside have in his career \\
NQ (English) & who migrated to the sahara savanna in present-day southeastern nigeria\\
 \bottomrule
\end{tabular} 
\end{center}\vspace*{-10pt}
\caption{Examples of questions we cannot find any web resources including answers.}
\label{tab:missing_info}
\end{table*}

\end{document}